\documentclass[journal]{IEEEtran}
\IEEEoverridecommandlockouts
\usepackage{cite}
\usepackage{amsmath,amssymb,amsfonts}
\usepackage{algorithmic}
\usepackage{graphicx}
\usepackage{textcomp}
\usepackage{xcolor}
\usepackage{microtype}
\usepackage{url}
\usepackage{wasysym}
\usepackage{enumitem}
\definecolor{codebg}{rgb}{0.95,0.95,0.95}

\def\BibTeX{{\rm B\kern-.05em{\sc i\kern-.025em b}\kern-.08em
    T\kern-.1667em\lower.7ex\hbox{E}\kern-.125emX}}

\begin{document}

\title{Human-Robot Interaction and Perceived Irrationality: A Study of Trust Dynamics and Error Acknowledgment\\
{\footnotesize \textsuperscript{}}
}
\author{Ponkoj Chandra Shill, Md Azizul Hakim, and Shawon Dey%
\thanks{Ponkoj Chandra Shill and Md Azizul Hakim are with the University of Nevada, Reno, NV 89557, USA. (e-mail: pshill@unr.edu, mdazizulh@unr.edu).}%
\thanks{Shawon Dey is with New Mexico State University, Las Cruces, NM 88003, USA (e-mail: sdey@nmsu.edu).}}

\maketitle


\begin{abstract}
As robots become increasingly integrated into various industries, understanding how humans respond to robotic failures and how these failures influence human trust in employing robots is critical. This study comprehensively examines trust dynamics and system design by analyzing human reactions to robot failures. We conducted a four-stage survey to explore how trust evolves throughout human-robot interactions. The first stage collected demographic data and initial trust levels. The second stage focused on preliminary expectations and perceptions of robotic capabilities. The third stage examined interaction details, including robot precision and error acknowledgment. Finally, the fourth stage assessed post-interaction perceptions, evaluating trust dynamics, forgiveness, and willingness to recommend robotic technologies. Results indicate that trust in robotic systems increased when robots acknowledged their errors or limitations. However, frequent or repeated failures led to a notable decline in trust. Participants also demonstrated greater willingness to recommend robots for future tasks when interactions involved a balance between transparency and accuracy, underscoring the importance of direct engagement in shaping trust dynamics. These findings provide valuable insights for designing more transparent, responsive, and trustworthy robotic systems. By advancing our understanding of human–robot interaction (HRI), this study helps guide the design of robotic technologies that build public trust and encourage broader adoption.
\end{abstract}

\begin{IEEEkeywords}
Human-Robot Interaction (HRI), Social Robotics, Error Acknowledgment, Trust Dynamics, ANOVA.
\end{IEEEkeywords}

\section{Introduction}
Recent advances in robotics, artificial intelligence, and Human-Robot Interaction (HRI) \cite{sheridan2016human,su2023} have collectively transformed robots from mere mechanical devices into intelligent agents capable of autonomous reasoning, social awareness, and adaptive decision-making. In contemporary settings, robots and autonomous systems are no longer confined to controlled laboratory settings but are increasingly integrated into diverse domains such as healthcare \cite{kyrarini2021survey}, manufacturing \cite{pedersen2016robot}, customer service, \cite{xiao2021robotics}, and automation \cite{dey2024distributed}, where they interact directly with humans in complex and dynamic environments. As robots increasingly assume roles that shape human decision-making and trust \cite{alex2021}, a critical question arises:

\begin{center}
    \textit{How do robot errors influence human trust?}
\end{center}

One of the key challenges in this context is understanding how humans perceive and respond to robot errors, especially in scenarios where robots are expected to deliver accurate information or guidance \cite{err-2017}. Such errors can result in a wide range of outcomes, from mistrust and frustration to potentially hazardous consequences depending on the operational context. Investigating human reactions to incorrect or misleading information provided by robots is not merely an academic pursuit but a critical step toward improving the design, reliability, and public acceptance of robotic systems \cite{frustration2021}.
\begin{figure}[htbp]
\centerline{\includegraphics[width=0.35\textwidth, height=0.35\textwidth]{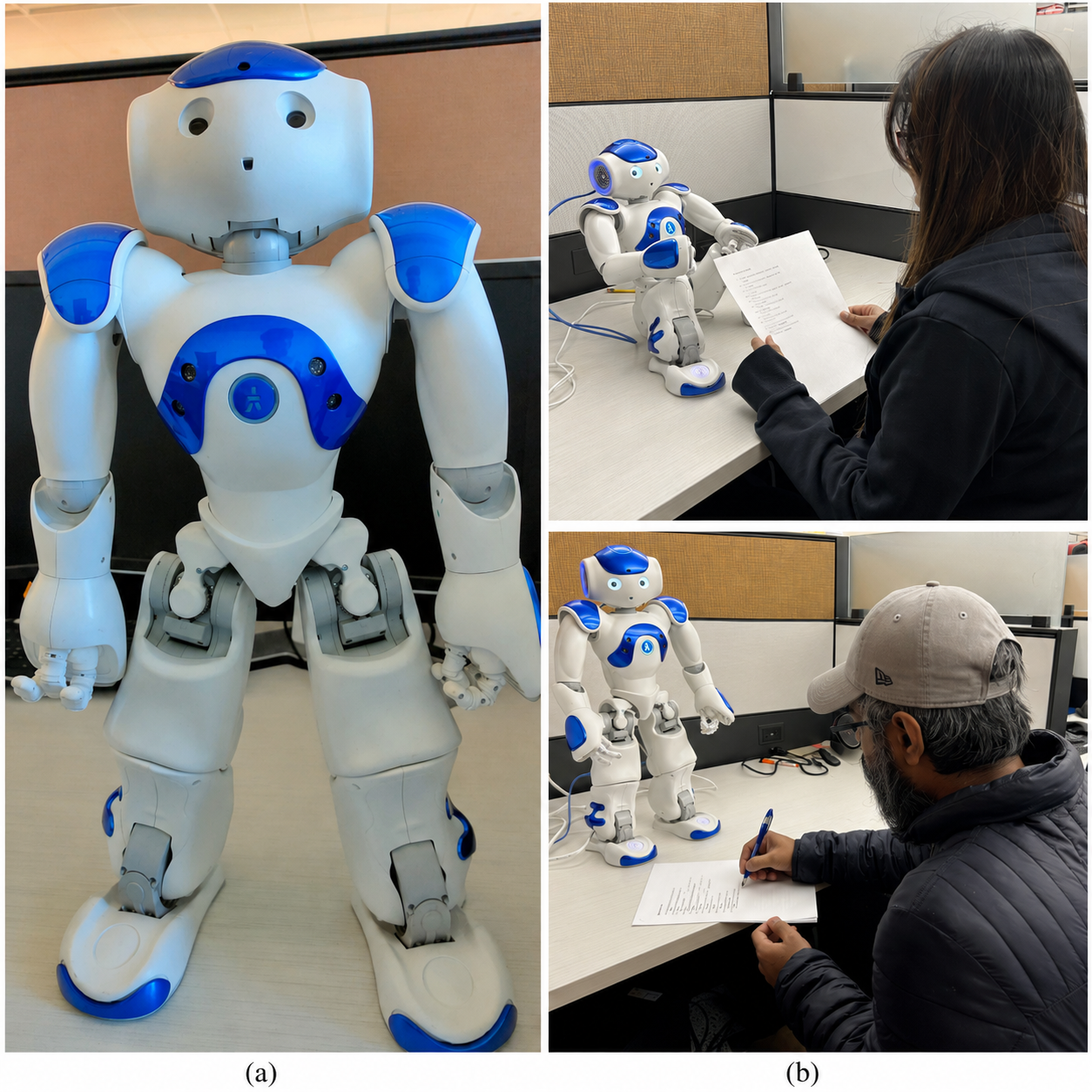}}
\caption{(a) NAO (b) Human–robot interaction scenario in which participants ask the robot a prescribed set of questions and provide feedback based on the responses received from the robot.}
\label{nao}
\end{figure}
Recent research has revealed multiple aspects of human-robot interaction and human responses to robot errors. D. Kontogiorgos et al. \cite{kontogiorgos2021crossCorpusAnalysis} highlighted the importance of multimodal cues in detecting and resolving conversational breakdowns in HRI, underscoring the complexity of human reactions and the crucial role of non-verbal signals in managing communication errors. Similarly, M. Stiber et al. \cite{stiber2022modeling} demonstrated that natural social responses can be effectively utilized for real-time detection and localization of robot errors in collaborative human-robot tasks.

Despite the growing body of research, a significant gap remains in understanding how robot mistakes and acknowledgments influence human trust. Gaining this understanding is vital for designing robots that promote transparency and reliability, enabling trustworthy and effective human-robot collaboration in real-world environments. This research aims to bridge this gap by examining human responses to robot errors and their impact on trust and system design in human-robot collaboration. Based on the study objectives, the following hypotheses are proposed:
\begin{itemize}
\item[H1:] When a robot acknowledges its errors or limitations, human trust and emotional acceptance toward the robot will increase.
\item[H2:] Humans are more tolerant of informational errors made by familiar agents (e.g., other humans or search engines) than by robots.
\end{itemize}

To test these hypotheses, the study employs a four-stage survey designed to capture key aspects of human-robot interaction. Through this structured investigation, this research seeks to deepen understanding of human-robot trust dynamics and provide insights for designing more reliable, user-centered, and context-aware robotic systems capable of responding empathetically to human reactions.
\begin{figure*}[htbp]
    \centering
    \includegraphics[width=.9\textwidth, height=0.29\textwidth]{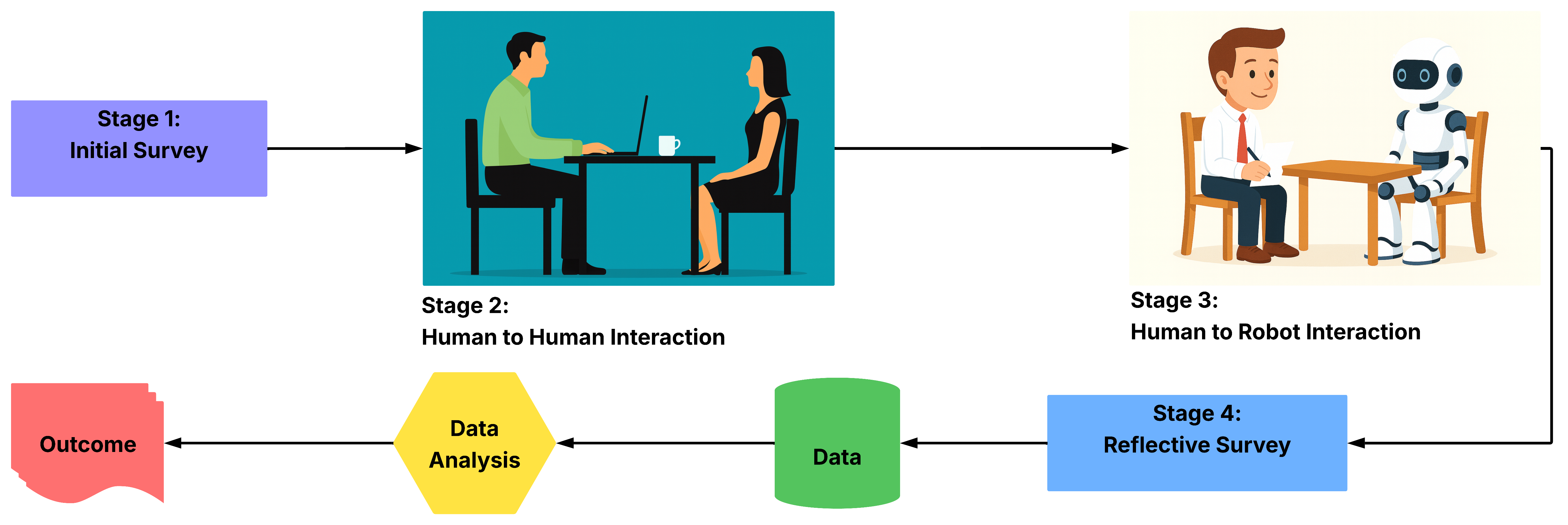}
    \caption{Research Methodology}
    \label{methodology}
\end{figure*}
\section{Related Work}
Human-robot interaction (HRI) involving robotic errors has been the focus of numerous studies exploring how humans perceive, interpret, and respond to such failures. Prior research has examined the psychological effects, behavioral adaptations, and multimodal cues associated with these interactions. This section reviews key findings from earlier work and highlights how the present study extends existing knowledge by addressing remaining gaps in the literature.

Previous studies have shown that human tolerance toward robot mistakes depends on the context and nature of interaction. For example, Yasuda and Matsumoto \cite{yasuda2013psychologicalImpact} found that robot errors can strongly influence human trust and perception of the robot. Similarly, Traeger et al. \cite{traeger2020robotsMistakesYaleNews} examined the role of robot error admission in human-robot conversations. Their study showed that when robots admit mistakes, the human-to-human interactions become more constructive and cooperative. Next, Salek Shahrezaie et al. \cite{SalekShahrezaie2020} explored the role of homophily, the tendency to connect with similar others-in human-robot interaction, showing that shared interests and perceived similarity can strengthen trust and emotional bonding between humans and robots. Hayes et al. \cite{Hayes2016} examined implicit human responses to robot mistakes in learning from demonstration scenarios, identifying nonverbal behaviors as key signals for building mutual understanding and improving feedback during interaction. Salem et al. \cite{salem2015trustInHRI} explored how faulty robot behavior affects trust and cooperation, revealing that although performance errors reduce perceived reliability, they do not always weaken people’s willingness to collaborate with the robot. Stiber and Huang \cite{stiber2021} further analyzed human reactions to robot errors of varying severity, showing that more severe errors evoke faster and stronger social responses, offering valuable insights for early error detection and adaptive robot behavior.


Our research extends these foundational studies by examining how error acknowledgment, apology behaviors, and the frequency or number of errors influence human trust, forgiveness, and willingness to engage in future interactions. While previous work has explored human reactions to robot errors, little attention has been given to how a robot’s self-awareness of mistakes, transparent communication, and error frequency affect the recovery of trust rather than its loss. We address this gap through a controlled, multi-stage study that compares human responses in human-human and human-robot interactions under similar error conditions. By analyzing both verbal and nonverbal reactions, our study provides new insights into how transparency, apology, and error frequency collectively shape trust dynamics in task oriented environments. These findings contribute to the development of more adaptive, socially aware robotic systems capable of maintaining and rebuilding trust through effective communication and accountability.

\section{Experimental Setup}
\textbf{System Configuration:} In this experiment, the NAO robot~\cite{nao}, a humanoid platform renowned for its interactive capabilities, was employed as illustrated in Fig.~\ref{nao}. NAO was chosen for this study due to its ability to facilitate natural and engaging human-robot conversations, making it an ideal platform for examining trust dynamics in human-robot interaction (HRI). Equipped with speech recognition, natural language processing, and expressive gestures, the robot can engage participants in dialogues that closely resemble human conversation. During the experiment, NAO was positioned directly in front of participants in a controlled laboratory environment. Each session began with a greeting from the robot, followed by an invitation for participants to ask questions from a predetermined set. NAO’s hardware, featuring advanced sensors, a 1.91 GHz Quad-Core Atom E3845 processor, and 4 GB of DDR3 RAM, enabled real-time data acquisition and efficient processing of multimodal interactions.

\begin{figure*}[htbp]
    \centering    \includegraphics[width=1\textwidth]{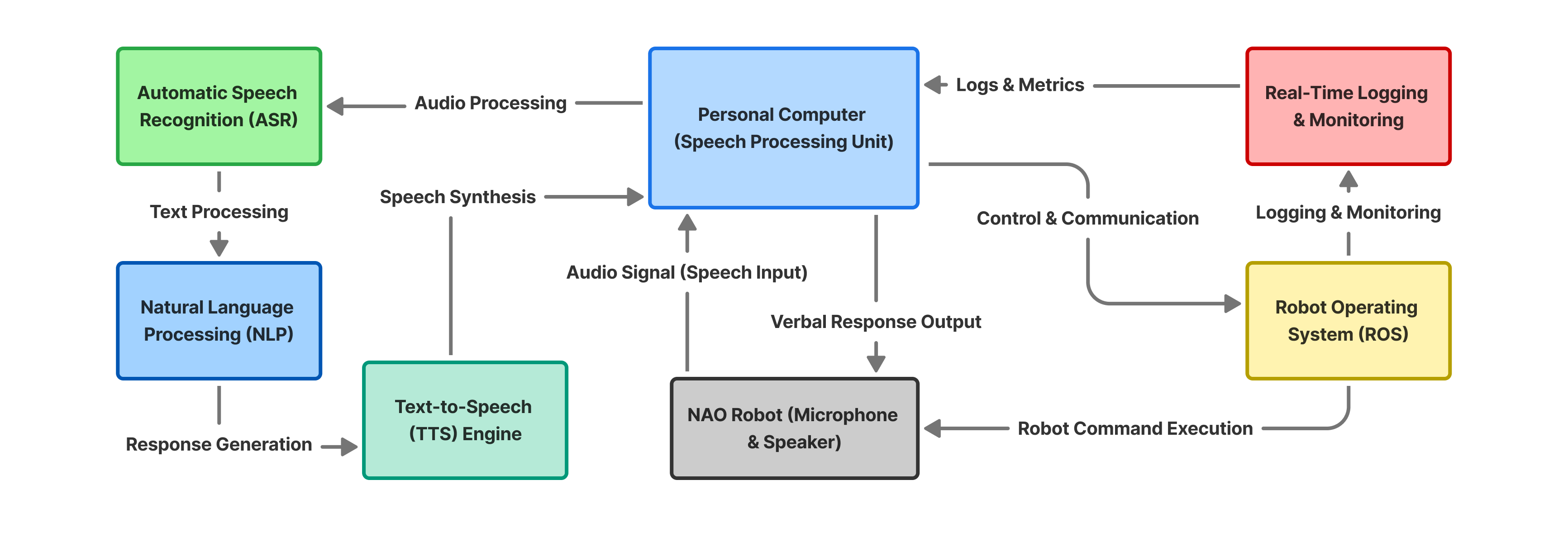}
    \caption{Hardware and software integration of the NAO robot, illustrating the interaction flow between the robot, external processor, and speech recognition modules.}
    \label{hardware_software_integration}
\end{figure*}

\textbf{Hardware and Software Integration: } The NAO robot was connected to a personal computer that served as the primary unit for speech recognition and response generation. This configuration utilized the Robot Operating System (ROS) to manage modular control, real-time data processing, and communication between NAO’s onboard hardware and external computing resources. Participant speech inputs were captured through NAO’s multidirectional microphone array and transmitted to the external processor via a low-latency communication link. An Automatic Speech Recognition (ASR) engine converted the audio to text, which was then processed by a Natural Language Processing (NLP) pipeline for intent recognition, contextual understanding, and response generation. The system’s text-to-speech (TTS) module synthesized these outputs, allowing NAO to deliver natural, human-like verbal responses in real time. ROS middleware also supported continuous monitoring and data logging of interactions, facilitating system debugging and performance analysis. Fig.~\ref{hardware_software_integration} depicts the integrated hardware–software architecture, highlighting the interaction between NAO, the external processor, and the speech recognition system.

\textbf{Speech Recognition:} The speech recognition system employed the Python SpeechRecognition library to transcribe spoken input into text. Using the Natural Language Toolkit (NLTK), the text was then processed through tokenization, lemmatization, and stop-word filtering to enhance contextual understanding. The NLP module analyzed each transcribed query to detect keywords corresponding to one of ten predefined questions. Once a match was identified, a relevant response was generated, converted to text, and transmitted to the NAO robot. NAO then vocalized the response through its built-in speakers using text-to-speech synthesis.

\textbf{Error Induction and Trust Evaluation Protocol:} To examine the effects of robot errors on human trust and perception, the experimental design incorporated a controlled error-generation mechanism. A randomized algorithm intermittently selected specific responses to be intentionally incorrect, simulating real-world misinterpretations or processing failures. In one setup, a fixed number of errors was maintained, while the randomizer determined which questions would trigger them. In another, participants experienced varying numbers of preselected errors, from low to high, to evaluate how increasing error frequency influenced trust dynamics. When an incorrect response occurred, the system automatically detected it and initiated an apology sequence. The NAO robot verbally acknowledged its mistake, expressed regret, and, in some cases, provided the corrected answer. This mechanism enabled the study of both trust degradation and recovery following robotic failures. After completing the ten-question interaction, the NAO robot thanked the participant, concluding the session. This experimental framework provided valuable insights into how error frequency and apology behavior shape human trust and acceptance in HRI.
\section{Methodology}
The research methodology was designed to assess participants’ immediate emotional responses and subsequent reflections after interacting with the robot, which is depicted in Fig.~\ref{methodology}. A mixed-methods approach was employed, integrating qualitative and quantitative data through a structured four-stage survey.
\subsection{Four Stage Study}
\subsubsection{Preliminary Survey} This stage collected participants' demographic information and also the baseline attitudes toward robots. Conducted before the interaction, the survey gathered participants’ background and preconceptions about robots and automation. It included questions on age group, gender, country, prior interaction with robots, perception of robots, and trust in automation. The results established a baseline for understanding each participant’s initial stance, serving as a reference for analyzing changes in trust and perception after the human-robot interaction.
\subsubsection{Perception of Interaction with Humans} In this interaction stage, each participant individually engages with a human interlocutor and asks a predefined set of ten general knowledge questions. The human intentionally provides a few randomly selected incorrect answers to simulate error occurrences, each followed by an acknowledgment and an apology. Upon completion of the interaction, participants complete a post-interaction survey evaluating the accuracy of the human’s responses and documenting their immediate reactions and perceptions regarding the human’s performance and error-handling behavior.
\subsubsection{Perception of Interaction with Robot} In the robot interaction stage, each participant individually interacts with NAO, a humanoid robot programmed to exhibit controlled errors, using the same predefined set of ten general knowledge questions. The study consists of two stages. In the first stage, NAO provides a fixed number of randomly selected incorrect answers while correctly answering the remaining questions. After each error, the robot acknowledges the mistake and issues an apology. In the second stage, the number of incorrect responses varies randomly across participants to examine how the frequency of errors influences their perceptions of the robot, building on the acknowledgment and apology mechanism from the first stage. Following each interaction, participants complete a post-interaction survey to report the number of questions asked, assess the accuracy of NAO’s responses, note the observed errors, and indicate whether the robot apologized. This stage is crucial for capturing participants’ immediate reactions and perceptions of NAO’s performance, accuracy, and error-handling behavior, providing key insights into human-robot interaction dynamics.
\subsubsection{Reflective Survey} This stage evaluates changes in participants’ perceptions and trust levels following their interactions with both the human and NAO. It serves as a key component of the methodology for collecting detailed post-interaction survey data, focusing on participants’ nuanced reactions and attitudes. Specifically, it examines several dimensions, including the impact of errors on trust, the effect of unacknowledged errors, comparative forgiveness between human and robot, the effectiveness of the robot’s apology, the importance of error acknowledgment, willingness to rely on the robot after an error, intention for future interaction, perceived reliability, and likelihood of recommendation. Overall, this stage aims to capture the depth and scope of human responses to robotic performance, offering valuable insights into trust formation, reliability assessment, and the psychological implications of robotic errors and their acknowledgment.
\subsection{Participant Selection}
\begin{figure}[t]
    \centering
    \includegraphics[width=\columnwidth]{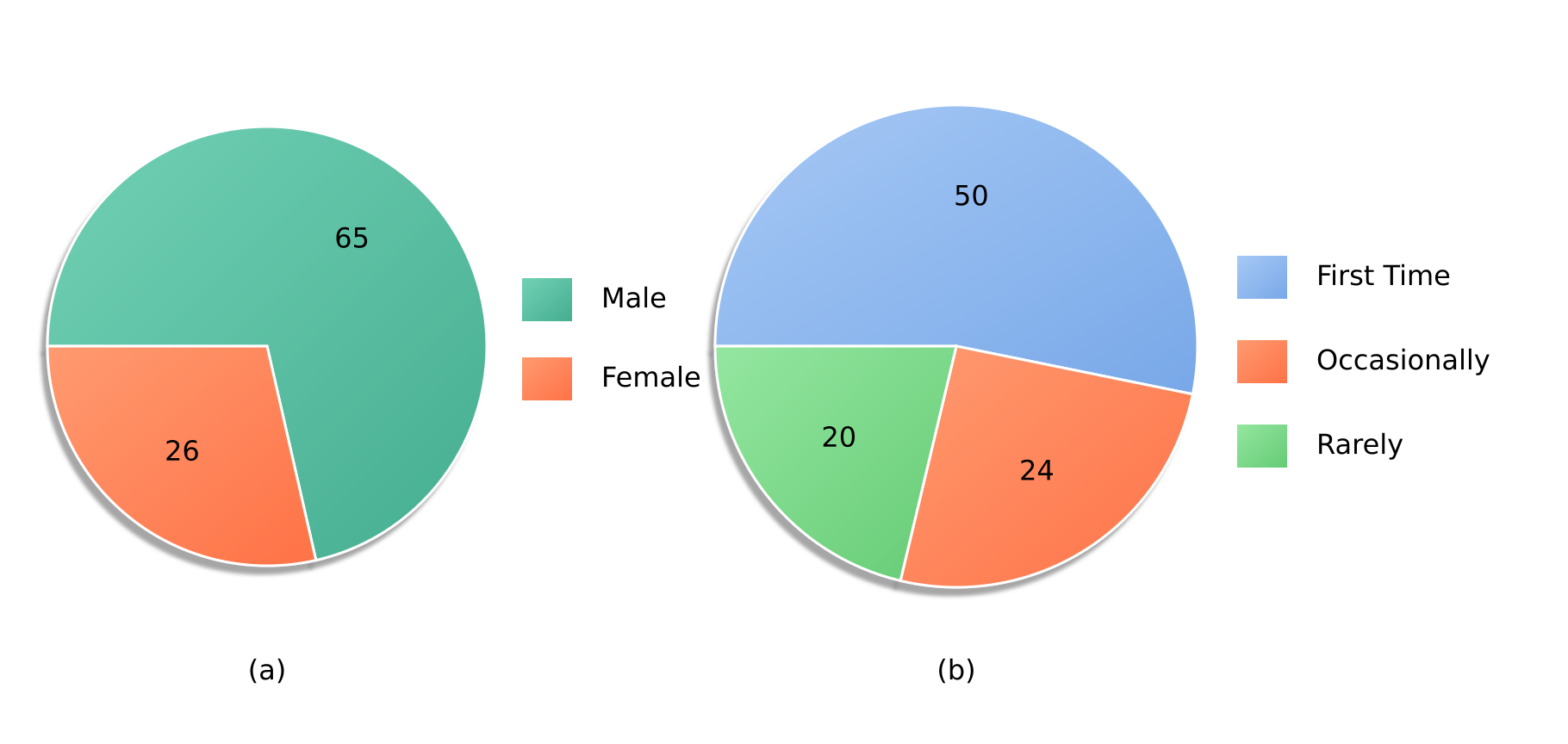}
    \caption{Participant characteristics: (a) gender distribution and (b) prior robot experience.}
    \label{participation}
\end{figure}

The participant group for this study consists of 94 individuals from various countries, aged between 21 and 64 years old, with 29 females and 65 males. Most participants (50 out of 94) have no prior experience in interacting with robots, making this study their first significant exposure to such technology (see Fig.~\ref{participation}). Generally, their perception of robots tends to lean towards viewing them as "somewhat reliable and useful". However, there is variability in their trust levels towards automation technology, with a moderate level of trust being the most common trend observed.
These insights indicate that while the participant group is diverse, there exists an overall optimism about the potential of robotic systems despite limited prior exposure. Moreover, they demonstrate openness to advancements in human-robot interactions (HRI), highlighting the importance of establishing trust in automation for broader acceptance of future technologies.
The findings underscore the necessity of developing user-friendly interfaces and ensuring transparency in robot functionalities, as these factors could significantly enhance perceptions and trust in such technologies.

\subsection{Questions Selection}
To assess the robot’s ability to answer factual questions, ten general knowledge questions were selected, covering diverse topics such as history, geography, science, mathematics, language, and symbolic recognition. Each question has a clear, objective answer, enabling a straightforward evaluation of the robot’s accuracy. The selected questions are as follows:
\begin{enumerate}
    \item What is the date of Independence Day in the USA?
    \item How many stars are on the flag of the USA?
    \item What is the capital of France?
    \item What is the name of the largest planet in our solar system?
    \item What is the currency of the USA?
    \item How many sides does a triangle have?
    \item What is the opposite of ‘day’?
    \item What shape is a ‘stop’ sign?
    \item Who was the first president of the USA?
    \item Who wrote the play \textit{Romeo \& Juliet}?
\end{enumerate}

These questions serve multiple purposes. First, they assess the robot’s factual recall ability, ensuring it can deliver accurate and well-structured responses to common knowledge queries. Second, questions such as number $8$ (stop sign shape) evaluate the robot’s capacity to recognize real-world symbols, an essential skill for navigation and contextual awareness in human-robot interaction. Additionally, questions such as number $7$ (opposite of “day”) test basic language comprehension and semantic understanding.
\section{Results and Discussion}

\subsection{Trust and Forgiveness in Human-Robot Interaction}

The survey data provides valuable insights into human reactions to robot interactions, shedding light on several key aspects:

\begin{itemize}
\item \textbf{Perceived Reliability and Usefulness:} Participants with limited prior experience with robots generally found them somewhat reliable and useful. Trust levels varied significantly, with 58\% expressing moderate trust before interacting with the robot. However, some participants remained skeptical, suggesting that initial trust in automation is influenced by personal experience and exposure.

\item \textbf{Reaction to Errors:} All participants noticed errors in the robot’s responses. However, 84\% reported increased trust when the robot acknowledged and apologized for its mistakes. This underscores the importance of transparency in human-robot interaction, as error acknowledgment appears to enhance trust rather than diminish it.

\item \textbf{Willingness to Rely on Robots:} Despite observing mistakes, many participants expressed confidence in using robots for future tasks. This suggests a growing acceptance of robots, where users recognize that even intelligent systems can make errors just like humans. Designing robots that communicate their limitations effectively could further strengthen their perceived reliability.

\item \textbf{Trust Post-Mistakes:} After witnessing errors, trust in the robot either remained unchanged or increased for most participants. This indicates that error-handling strategies (such as apologies and explanations) play a crucial role in trust dynamics. The results suggest that when robots acknowledge their errors, users are more likely to maintain or even increase their trust in automation.
\item \textbf{Forgiveness of Human Errors:}
Humans are generally more forgiving of mistakes made by other humans compared to those made by robots. This is largely due to social and cognitive biases, where human errors are often attributed to situational factors such as distraction, fatigue, or momentary lapses in attention. In contrast, robotic errors are more likely to be perceived as system failures, raising concerns about reliability and competence. This tendency aligns with our findings, where participants were significantly less forgiving towards robots than humans or search engines. On the other hand, some people are neutral. These results suggest that if robots incorporate more human-like transparency, social cues, and adaptive learning behaviours, they may be perceived as more trustworthy and acceptable in real-world applications.
\end{itemize}

\begin{figure}[t]
    \centering
    \includegraphics[width=\columnwidth]{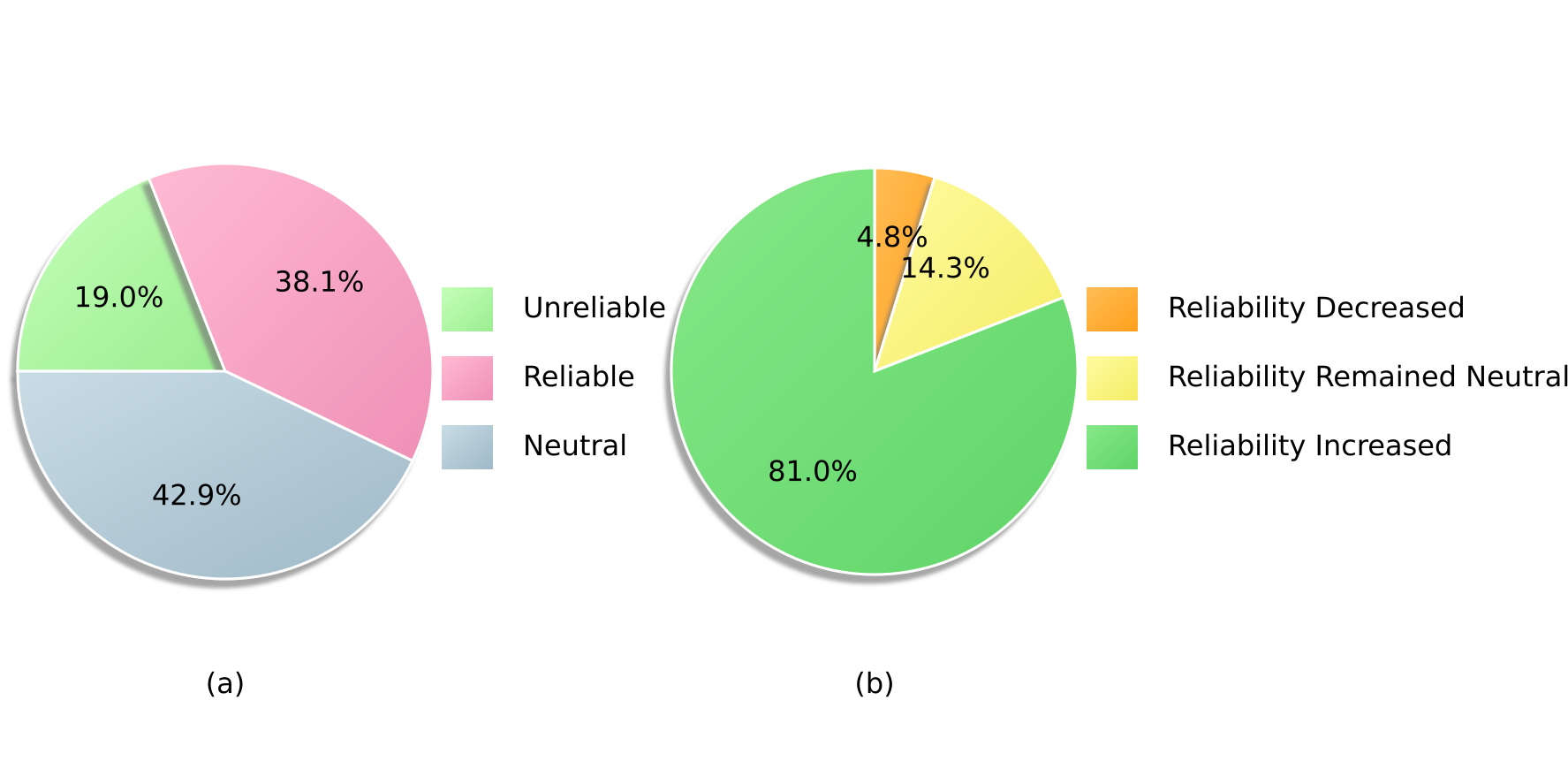}
    \caption{Perception of robot reliability: (a) before the interaction study and (b) after the interaction study.}
    \label{comparison}
\end{figure}

\begin{figure}[htbp]
\centering
\includegraphics[width=.45\textwidth, height=0.26\textwidth]{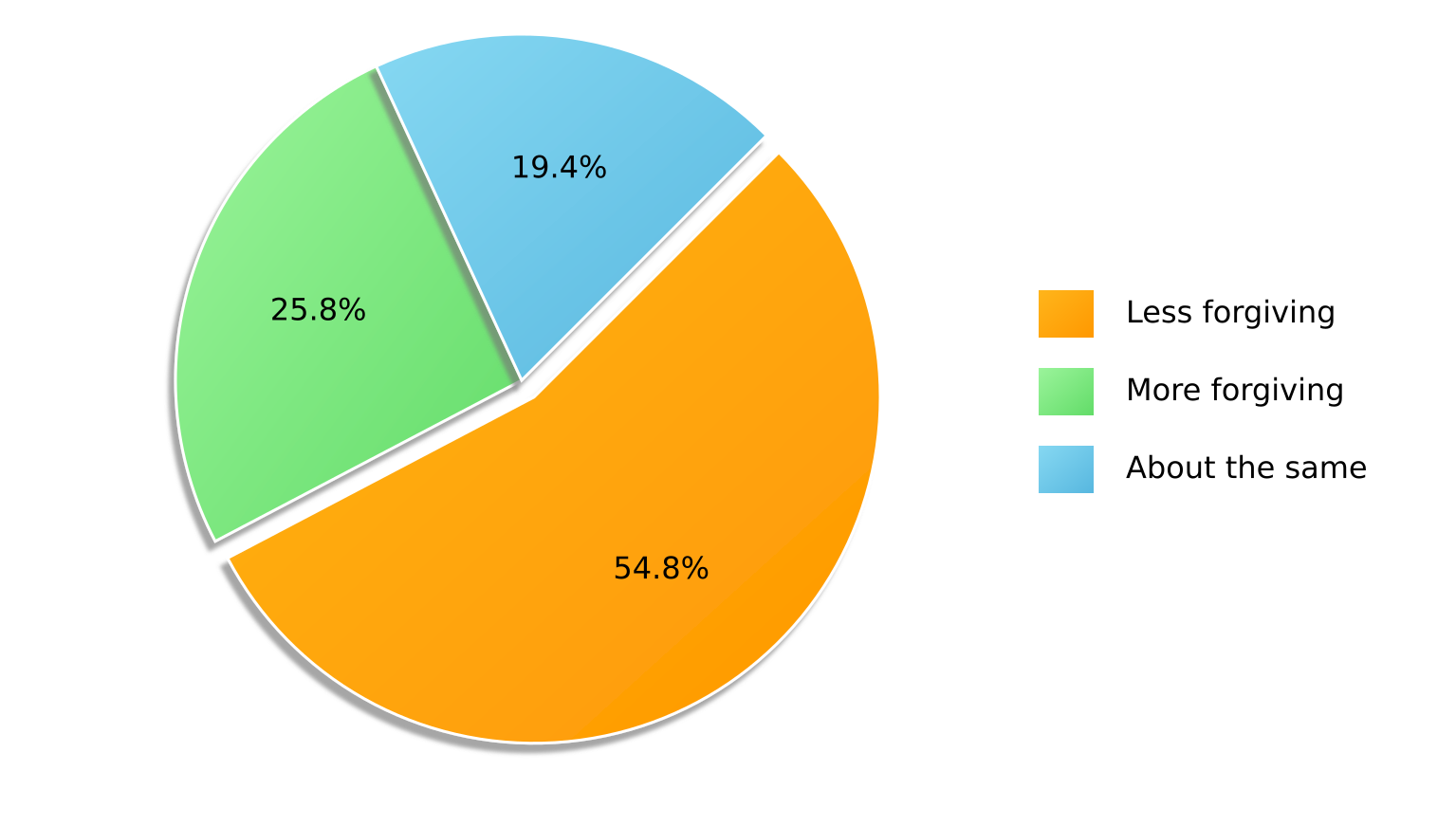}
\caption{Forgiveness towards human vs robots. People are more forgiving to humans than robots.}
\label{forgiveness}
\end{figure}

\subsection{Evaluation of Hypotheses}

\subsubsection{Support for H1: The impact of error acknowledgment on trust}
To evaluate H1, all participants were exposed to errors in the robot’s responses. Interestingly, 84\% of participants reported increased trust when the robot acknowledged its errors and apologized, highlighting the importance of transparent communication in fostering user trust.

A comparative analysis of participants' trust levels before and after the study further reinforces this finding. As shown in Fig.~\ref{comparison}, only 32.3\% of participants initially perceived the robot as reliable, with 19.4\% considering it unreliable and 48.4\% remaining neutral. However, after interacting with the robot, 83.9\% of participants reported increased reliability perception, while only 6.5\% experienced a decrease in trust. Additionally, 84\% of participants expressed a willingness to recommend the use of robots in high-stakes domains. These results strongly support H1, demonstrating that when robots acknowledge their limitations and errors, human trust and emotional acceptance significantly increase.

\subsubsection{Support for H2: Forgiveness of errors from different sources}
To assess H2, participants were asked to compare their level of forgiveness toward informational errors made by humans, robots, and internet search engines. The results indicate that humans are generally more forgiving of errors from familiar sources compared to robots.

As illustrated in Fig.~\ref{forgiveness}, 54.8\% of participants were less forgiving toward robots, whereas 25.8\% were more forgiving toward robots than humans, and 19.4\% expressed equal levels of forgiveness for both sources. These findings suggest that people hold robots to a higher standard of accuracy than human or internet-based sources, possibly due to differing expectations regarding machine intelligence and perceived competence. These findings provide strong empirical support for H2, confirming that humans are more likely to overlook errors when they come from familiar sources rather than from robots.

\subsection{Statistical Analysis of Trust Level Changes}
\begin{figure*}[htbp]
    \centering
    \includegraphics[width=1\textwidth]
    {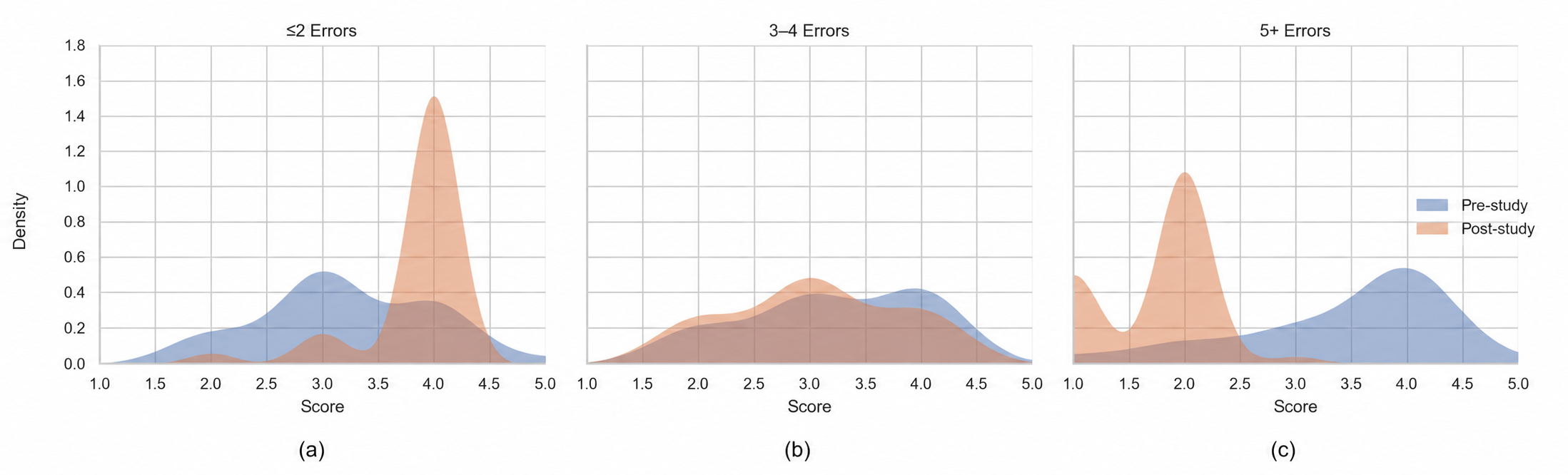}
    \caption{Density distributions of participants' trust scores before and after the experiment, grouped by the number of robot errors: (a) two or fewer errors, showing post-study scores concentrated near higher trust levels; (b) three to four errors, showing substantial overlap between the pre-study and post-study distributions; and (c) five or more errors, showing post-study scores shifted toward lower trust levels. Blue represents pre-study responses, and orange represents post-study responses.}
    \label{trustlevel}
\end{figure*}

To provide robust experimental evidence, we conducted a one-way ANOVA test \cite{anova} to analyze the change in participants’ trust levels before and after the experiment. The results, visualized in Fig.~\ref{trustlevel}, reveal a highly significant increase in mean trust levels post-experiment.

As seen in Fig.~\ref{trustlevel}, the density distribution before the experiment (blue) shows a broader spread with a lower peak, indicating a more varied and generally lower level of trust in the robot. In contrast, post-experiment (red), the distribution becomes narrower and sharply centered around a higher trust level, suggesting a more consistent and elevated perception of reliability. The one-way ANOVA test confirms a statistically significant overall change in trust levels across experimental groups, yielding an F-statistic of 150.69 and a p-value of 3.16 × 10\^{-13}. This highly significant result indicates that participant trust in the robot significantly increased after the experiment.


\begin{table}[htbp]
\centering
\small
\caption{Contingency Table for Trust Levels Before and After Experiment}
\begin{tabular}{|p{1.5cm}|p{1.5cm}|p{1.5cm}|p{1.5cm}|}
\hline
\multicolumn{1}{|c|}{} & \multicolumn{3}{c|}{\textbf{Trust/Reliability Level After Experiment}} \\ \hline
\textbf{Trust Level Before Experiment} & \textbf{Increases reliability} & \textbf{No change} & \textbf{Decreases reliability} \\ \hline
Completely trusted (a) & 7 & 0 & 0 \\ \hline
Moderately trusted (b) & 0 & 11 & 0 \\ \hline
Slightly trusted (c) & 0 & 0 & 1 \\ \hline
\end{tabular}
\label{tab:contingency}
\end{table}

To compare the trust levels before and after the experiment, we created a contingency table. Table~\ref{tab:contingency} allows us to visualize the relationship between the trust level before the experiment (rows) and the trust/reliability level after the experiment (columns). Based on the table, we can confidently say that the trust level increased after the experiment, indicating that the participants perceived the robot as more reliable following the experimental intervention. The chi-square test has been conducted on the contingency table, yielding the following results:
\begin{itemize}
    \item Chi-Square Statistic: 38.0
    \item p-value: \(1.12 \times 10^{-7}\)
    \item Degrees of Freedom: 4
    \item Expected Frequencies:
\end{itemize}

\[\begin{bmatrix}2.579 & 4.053 & 0.368 \\4.053 & 6.368 & 0.579 \\0.368 & 0.579 & 0.053 \\\end{bmatrix}\]

The chi-square analysis ($\chi^2 = 38.0$, $p = 1.12 \times 10^{-7}$) confirms a significant association between pre-experiment and post-experiment trust levels. This result validates that the observed increase in trust was not random but statistically meaningful, reinforcing the effectiveness of error acknowledgment in human-robot trust building.


\subsection{User Perspectives on Trust and Errors}
This research provides additional insights into key aspects of human-robot interaction based on participant responses.
\\

\subsubsection{Trust Impact of Unacknowledged Errors}
This research indicates that 77\% of participants experienced a decrease in trust when errors were not acknowledged. This finding underscores the importance of transparent communication in maintaining user trust in robotic systems. When a robot failed to acknowledge its mistakes, participants perceived it as less reliable, reinforcing the idea that users expect robots to be accountable for their errors.

\subsubsection{Importance of Acknowledging Errors}
All participants emphasized that robots admitting mistakes is crucial for trust maintenance. Acknowledgment of errors not only enhances transparency but also reassures users that the system is aware of its limitations. These findings suggest that designing robots with socially appropriate error-handling mechanisms can improve their perceived reliability and overall trustworthiness.

\subsubsection{Effectiveness of Robot's Apology}
87\% of participants found that a robot’s apology after making a mistake was effective in restoring trust. This result suggests that apologetic behaviors can mitigate negative perceptions of robotic errors, making robots more acceptable in real-world applications. The ability of robots to acknowledge mistakes and issue apologies helps foster a more human-like interaction, reducing skepticism and reinforcing user confidence.

\subsubsection{Error Impact on Trust}
Errors had a notable impact on trust, with 62\% of participants reporting a negative effect on their perception of the robot’s reliability. However, this impact was not uniform across all participants. While some individuals were more resilient to errors, others became more skeptical of the robot’s capabilities. This variation suggests that users’ prior experiences and expectations of automation influence how they perceive robotic reliability.


\subsubsection{Perceived Reliability of Robots}
Despite concerns about errors, 84\% of participants regarded robots as reliable after interacting with them. This aligns with the initial expectation that robots, though imperfect, are still functional and useful. The findings suggest that robot reliability is judged not just by error occurrence but also by how the robot handles mistakes.

\subsubsection{Future Interaction Intent}
93\% of participants expressed willingness to engage with robots in future tasks, demonstrating a positive outlook on human-robot interaction. Despite encountering errors, participants remained open to using robotic systems, suggesting that trust in automation can be strengthened through effective error-handling strategies and transparent communication. The findings highlight the importance of designing robots that not only perform tasks efficiently but also manage errors in a human-like and socially acceptable manner.

\section{Conclusion and Future Work}

In this study, we found that participants’ confidence in robots varied, with many exhibiting some degree of trust, even among those who had never interacted with robots before. Our experimental results support both hypotheses (H1 and H2).
Regarding H1, we observed that trust in robots was influenced by whether the robot acknowledged its mistakes. The majority of participants recognized the robot’s limitations, and 84\% reported increased trust when errors were admitted and apologized for. Despite being aware of errors, participants remained willing to interact with robots in future tasks, highlighting the importance of transparent error acknowledgment in trust-building.
For H2, our results confirm that humans are generally more forgiving of errors made by other humans or search engines than those made by robots. 55\% of participants were less forgiving of robotic mistakes, suggesting that people hold robots to higher accuracy standards than they do humans. However, when robots admitted their mistakes, trust improved, indicating that error-handling mechanisms play a crucial role in shaping user perceptions of robotic reliability.
Interestingly, even when mistakes were made, participants often perceived robots as equally or more trustworthy, suggesting that public acceptance of robots is increasing. Our findings demonstrate that people’s trust in robots is nuanced, balancing skepticism with an appreciation of their potential benefits.

For future research, we aim to investigate how long-term trust and reliance on robots are influenced by the type and frequency of errors they make. We will also examine how individuals from diverse demographic and cultural backgrounds respond to robot interactions. Additionally, to improve robot dependability and user trust, we will work on enhancing their ability to recognize and correct mistakes more effectively.
Furthermore, we plan to conduct similar experiments with different types of robots to analyze how people react to various robotic designs and to assess how different robot types influence human trust in automation. These studies will provide deeper insights into the social and cognitive factors shaping trust in robotic systems.
While these questions focus on fact-based knowledge, future work may expand the scope to include multi-step reasoning tasks and handling of ambiguous questions. This would provide a more comprehensive evaluation of the robot’s cognitive processing and adaptability in complex conversational settings.

\section*{ACKNOWLEDGMENTS}
We appreciate all of the help that was provided by Dr. David Feil-Seifer for sharing their opinions with us. We also appreciate the participants who helped us by participating in our survey.

\bibliographystyle{ieeetr}
\bibliography{roboticsreferences}

\clearpage 

\onecolumn


\end{document}